\documentclass[conference]{IEEEtran}

\usepackage[letterpaper,top=1.9cm,bottom=2.54cm,left=1.57cm,right=1.57cm]{geometry}
\usepackage{times}
\usepackage{cite}
\usepackage{amsmath,amssymb,amsfonts}
\usepackage{algorithmic}
\usepackage{graphicx}
\usepackage{textcomp}
\usepackage{xcolor}
\usepackage{setspace}
\usepackage{caption}
\usepackage{epsfig}
\usepackage{multicol,multirow}
\usepackage{booktabs}
\usepackage{verbatim}
\usepackage{float}
\usepackage{adjustbox}
\usepackage{mathtools}
\usepackage{cite}
\def\BibTeX{{\rm B\kern-.05em{\sc i\kern-.025em b}\kern-.08em
    T\kern-.1667em\lower.7ex\hbox{E}\kern-.125emX}}
\begin{document}

\title{LFA-Net: A Lightweight Network with LiteFusion Attention for Retinal Vessel Segmentation}

\author{
     \IEEEauthorblockN{Mehwish Mehmood \textsuperscript{*}\footnotemark} 
    \IEEEauthorblockA{
          \textit{School of EEECS}\\
        Queen's University Belfast, \\
        Belfast, BT7 1NN, United Kingdom \\
        \texttt{mmehmood01@qub.ac.uk}
    }

    \and
    \IEEEauthorblockN{Ivor Spence} 
    \IEEEauthorblockA{
          \textit{School of EEECS}\\
        Queen's University Belfast, \\
        Belfast, BT7 1NN, United Kingdom \\
        \texttt{i.spence@qub.ac.uk}
    }
    \and
    \IEEEauthorblockN{Muhammad Fahim} 
      \textit{School of EEECS} \\
        Queen's University Belfast, \\
        Belfast, BT7 1NN, United Kingdom \\
        \texttt{m.fahim@qub.ac.uk}
        
    }

\maketitle
\renewcommand{\thefootnote}{\fnsymbol{footnote}}  
\footnotetext{\textsuperscript{*} Corresponding author.}
\begin{abstract}
Lightweight retinal vessel segmentation is important for the early diagnosis of vision-threatening and systemic diseases, especially in a real-world clinical environment with limited computational resources. Although segmentation methods based on deep learning are improving, existing models are still facing challenges of small vessel segmentation and high computational costs. To address these challenges, we proposed a new vascular segmentation network, LFA-Net, which incorporates a newly designed attention module, LiteFusion-Attention. This attention module incorporates residual learning connections, Vision Mamba-inspired dynamics, and modulation-based attention, enabling the model to capture local and global context efficiently and in a lightweight manner. LFA-Net offers high performance with 0.11 million parameters, 0.42 MB memory size, and 4.46 GFLOPs, which make it ideal for resource-constrained environments. We validated our proposed model on DRIVE, STARE, and CHASE\_DB with outstanding performance in terms of dice scores of 83.28, 87.44, and 84.50\% and jaccard indices of 72.85, 79.31, and 74.70\%, respectively. The code of LFA-Net is available online https://github.com/Mehwish4593/LFA-Net.\\
\end{abstract}


\section{Introduction}
Blood vessel segmentation from retinal images is one of the primary tasks in analyzing the patients' ocular and systemic health. The structural changes in the vessels can indicate the presence of some pathology at early stages, such as diabetic retinopathy, hypertensive retinopathy, glaucoma, and cardiovascular or neurological diseases \cite{kapoor2022ai}. Automated vessel segmentation, however, remains a problematic task, as the complexity of the retinal vasculature structures and low-contrast boundaries make accurate segmentation challenging \cite{li2025adaptive}. Early deep learning solutions such as U-Net and its variants achieved significant improvements in vessel segmentation, but conventional CNNs struggle to capture long-range dependencies and fine details simultaneously \cite{tasker2025recurrent}. Real-world clinical settings, including mobile ophthalmology units, point-of-care diagnostics, or low-resource healthcare settings, lead to growing demands for segmentation models that are both diagnostically reliable and computationally efficient \cite{kumar2023fundus}. Attention included in compact architectures guarantees the preservation of fine details while improving feature representation without requiring the computational expense usually associated with transformer-based or deeper networks.\\

In this work, we propose LFA-Net, a lightweight segmentation model with a new vision-Mamba-inspired attention module called LiteFusion-Attention. We aim to achieve accurate segmentation while reducing the load for efficient deployment in real-time or low-resource clinical settings. The LiteFusion-Attention module is an effective attention mechanism for multiscale representation learning. It employs the idea of  modulation-based attention \cite{farooq2024lssf} for increased feature discrimination and uses residual connections to enhance the stable flow of gradients. Inspired by Vision Mamba \cite{zhu2024vision}, the module features dynamic token and channel mixing, capturing local and global dependencies.Furthermore, it applies global average and max pooling to enhance the context aggregation. The integration of these components in LiteFusion-Attention helps achieve an appropriate balance between the model complexity and performance, which is needed in resource-limited environments and real-time systems.\\

We trained and evaluated our model on three datasets: DRIVE, STARE, and CHASE\_DB, which are publicly available. The dice score, jaccard index, specificity, and sensitivity are computed to measure the performance of the parameters of our model. Also, the model's complexity is evaluated considering model size, trainable parameter count, and estimates in floating-point operations (FLOPs). The comparison of performance metrics with other competing models in advanced segmentation model practices validates the effectiveness of our model.\\

\begin{enumerate}
\item We propose LFA-Net, a novel lightweight retinal vessel segmentation network that balances high accuracy with computational efficiency, making it suitable for resource-constrained clinical environments.
    \item We design a multilevel attention mechanism, LiteFusion-Attention, specifically designed for retinal blood vessel segmentation, which combines residual learning, modulation-based attention, and dynamics inspired by Vision Mamba that performs context refinement and feature mixing across spatial and channel dimensions.
    \item We implement region-aware attention in selective skip connections to improve spatial context representation without increasing model complexity and present a thorough ablation providing architectural insights and proving that LFA-Net outperforms larger and more complex models, making it ideal for clinical deployment in the real world.
\end{enumerate}

\section{Related work}
Classic CNN architectures like U-Net apply encoder and decoder paths paired with skip connections to segment the retinal vessels. Recent works add custom attention and lightweight residual modules to strengthen these architectures. \\

Bhati et al. \cite{Bhati2025} proposed DyStA-RetNet, a shallow encoder-decoder CNN with attention blocks. This model integrates a multi-scale dynamic attention and a statistical spatial attention module into the encoder and a partial decoder to capture both high- and low-level information. This attention gating enhances segmentation by emphasizing vessels of different widths and drowning out background details. Abbasi et al. \cite{Abbasi2024} developed LMBiS-Net, a "Lightweight Multipath Bidirectional Skip" CNN tailored for retinal vessel segmentation. It features multipath feature extraction blocks to obtain vessel characteristics at different scales and bidirectional skip connections for the encoder and decoder to improve information exchange. LMBiS-Net is restricted to 0.172 million parameters while maintaining good accuracy. Shang et al.\cite{Shang2024} presented DCNet, which used a simpler framework for retinal vessel segmentation. The model is composed of only three convolutional layers; each layer has dilated kernels to capture multi-scale context, enabling spatially large receptive fields without extensive stacking. \\

A compact U-Net variation called FS-UNet is developed by Jiang et al. \cite{jiang2024retinal} for retinal vascular segmentation. Its goal is to minimize computing overhead while maintaining excellent segmentation accuracy. Self-attention algorithms and feature selection modules are integrated into the design to allow it to minimize background noise and concentrate on vessel-relevant features. S. Seo et al. introduced \cite{seo2025full} FSG-Net, which enhances U-Net by adding full-scale feature representations and an attention-guided filtering module that improves vessel segmentation by fine-tuning attention maps while using fewer parameters to maintain competitive performance. Amruthlal et al. \cite{qin2024branchfusionnet} proposed BranchFusionNet, which improves fine vessel detection without appreciably increasing model complexity by incorporating a Hybrid Residual Attention (HRA) module into a U-Net-like architecture. To increase the sensitivity of the model to small and low-contrast vessels, the HRA module integrates residual learning with spatial and channel attention. Li et al. \cite{li2020lightweight} developed LA-CNN, a Lightweight Attention Convolutional Neural Network for retinal vascular segmentation. They combined an attention module with a U-Net backbone to improve feature representation in the feature fusion process. While drastically lowering the number of parameters to 0.4M, the attention mechanism improves the delineation of thin and complicated vascular structures by capturing global contextual information.
\subsection{Transformer-Based and Hybrid Models}
To improve the global context modelling used in CNNs, self-attention and transformers have been integrated into image segmentation models \cite{badar2025transformer}. Vision Transformers, on their own, tend to be quite resource-intensive. Therefore, recent approaches developed hybrid architectures that integrate attention or transformer blocks with CNN backbones, which are lightweight and accurate.
Tong et al. \cite{Tong2024} developed a retinal vessel Segmentation Network (LiViT-Net), a U-Net with a MobileViT block embedded in every encoding stage. The MobileViT+ module efficiently merges local and global features. Local feature representations are further integrated with local feature representations through convolutional layers that capture fine spatial details, while Transformer blocks capture long-range dependencies. This design provides LiViT-Net with a strong balance between accuracy and model size. Mehmood et al. \cite{Mehmood2024} proposed a multi-task network, RetinaLiteNet, that systematically segments retinal vessels and the optic disc. The encoder contains standard convolutional layers and multi-head self-attention, which blend the outputs to exploit local detail and global context simultaneously. For the decoder, RetinaLiteNet employs skip connections enhanced by a Convolutional Block Attention Module (CBAM), which refines features in a spatial and channel-wise manner to modulate the features further.

\subsection{Vision Mamba-Inspired Models}

The application of state-space models (SSMs), as shown in the Vision Mamba framework, marks an exciting development in lightweight segmentation \cite{chrysos2025hadamard}. Mamba's sequence modelling layers have linear time scaling wherein long-range dependencies are captured. Recent models have particularly designed Vision Mamba for retinal vessel segmentation on fundus images, addressing issues with fine-grained structure recovery, vascular continuity, and computing efficiency.\\

Shao et al. \cite{shao2025tubular} presented TA-Mamba, a Tubular-Aware Mamba Block that respects the curved geometry and continuity of retinal capillaries through the use of high/low-frequency attention, tubular-aware gating, serpentine spatial convolution, and directional feature fusion. Ouyang et al. \cite{ouyang2025novel} suggested HREF-Net, a CNN–Mamba hybrid framework with a Dynamic Snake Visual State-Space (DSVSS) block. It combines dynamic snake convolution for curved vessel feature alignment with Mamba layers for global dependency modeling. By integrating multi-scale edge features, its companion Multi-scale Retina Edge Fusion (MREF) module improves border delineation. It demonstrates significant gains in boundary precision and low-contrast vessel detection on DRIVE, STARE, and CHASE\_DB. Serp-Mamba, proposed by Wang et al. \cite{wang2025serp}, examines retinal images with ultra-wide-field (UWF-SLO). To address class imbalance in high-resolution data, an Ambiguity-Driven Dual Recalibration (ADDR) module and a Serpentine Interwoven Adaptive (SIA) scan mechanism that simulates curved vessel traversal are presented. It is the first to apply Vision Mamba with SIA and ADDR to UWF fundus images. In order to improve vessel clarity and structure preservation, Wang et al. \cite{wang2025hybrid} developed MWRD incorporates Vision Mamba into a fundus image enhancement pipeline, which uses wavelet decomposition and reverse diffusion in addition to Mamba layers.\\

Compelling segmentation results were demonstrated by the authors, which indicates that the application of SSM-based blocks could enhance performance for thin, lengthy vessels that span wide image regions during retinal image segmentation.

\section{Methodology}
We introduce LFA-Net, a new lightweight model that includes multiscale feature extraction, region-aware attention (RAA), and the newly proposed LiteFusion-Attention module to effectively segment vascular structures and complexity through precise and robust segmentation. 
\begin{figure*}[!ht]
  \centering
    \includegraphics[width=\textwidth]{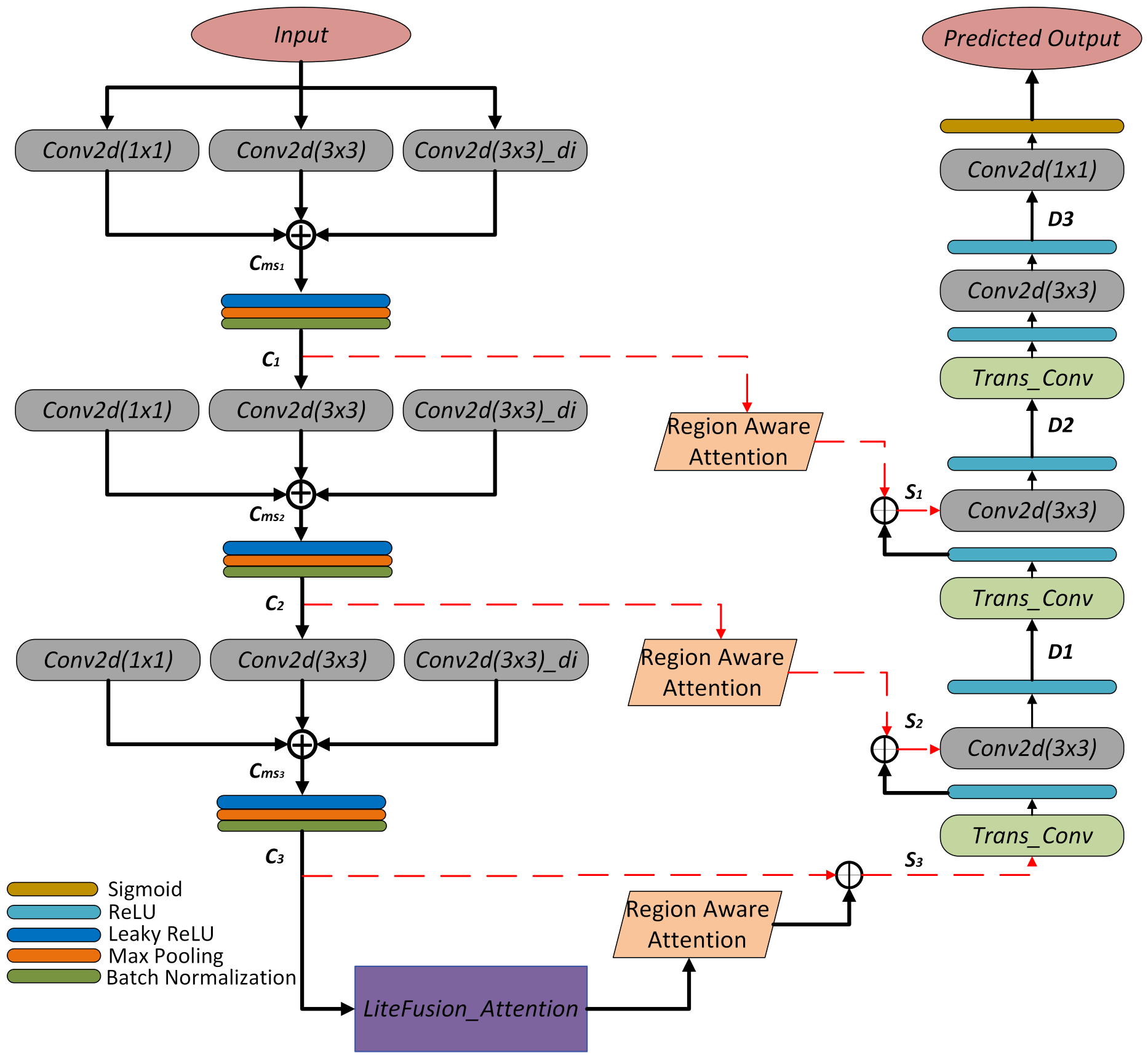}
    \caption{Architecture of the proposed LFA-Net}\label{LFA-Net}
\end{figure*}
\subsection{Model Architecture}

LFA-Net architecture is presented in Figure~\ref{LFA-Net}. It has three main components: multiscale convolution blocks for feature extraction, contextual attention refinement in selective skip connections with RAA \cite{naveed2024ra}, and a newly proposed LiteFusion Attention in the bottleneck for refined segmentation.

\begin{figure*}[!ht]
  \centering
    \includegraphics[width=\textwidth]{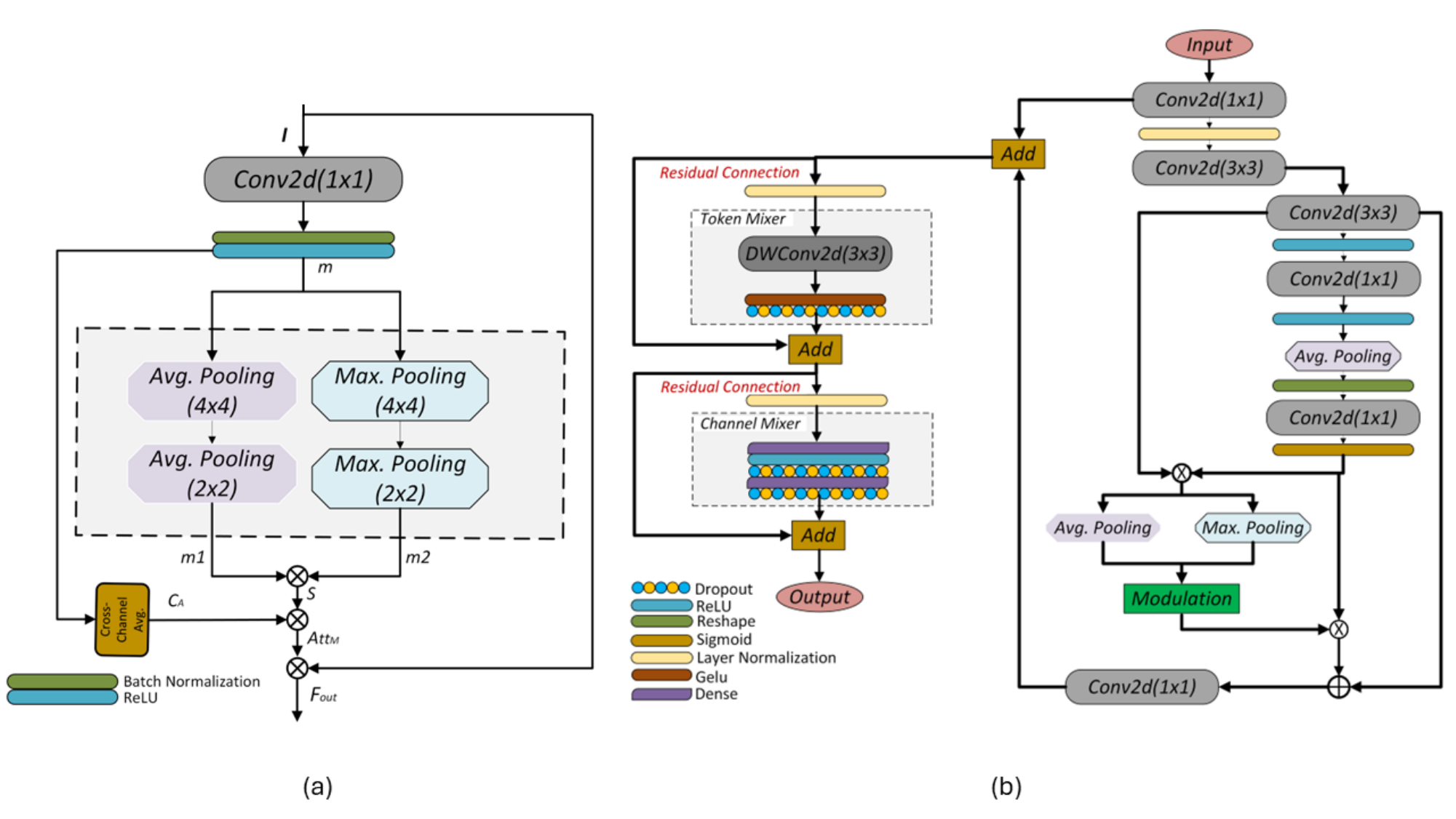}
    \caption{(a) Region Aware Attention Mechanism; (b) LiteFusion-Attention Module}\label{LiteFusion-Attention}
\end{figure*}

Each encoder block has three parallel convolution layers, $1\times1$, $3\times3$, and dilated $3\times3$ to capture multi-scale features with minimal parameters. The results are concatenated and passed through LeakyReLU activation, max pooling and batch normalization. 
\begin{equation}
  C_{ms_k} = C^{1\times1}(X_k) \oplus C^{3\times3}(X_k) \oplus C_{di}^{3\times3}(X_k)  \label{eq:encoder}
\end{equation}
\begin{equation}
   C_k = \mathit{LeakyReLU}(M_p(BN(C_{ms_k})))
  \label{eq:encoder}
\end{equation}
where $X_k$ is the input to stage $k$, $M_p$ is max pooling, $BN$ is batch normalization, and $\oplus$ denotes concatenation.\\

The bottleneck integrates LiteFusion-Attention ($\mathcal{F}_{lite}$) and RAA ($\mathcal{R}$) modules:
\begin{equation}
  S_3 = \mathcal{R}(\mathcal{F}_{lite}(C_3)) \oplus C_3.
  \label{eq:bottleneck}
\end{equation}

The decoder upsamples features via transposed convolutions and RAA-refined skip connections:
\begin{equation}
  S_2 = \mathcal{R}(C_2) \oplus \mathit{ReLU}(\text{TransConv}(S_3))
\end{equation}
\begin{equation}
  S_1 = \mathcal{R}(C_1) \oplus \mathit{ReLU}(\text{TransConv}(D_1))
\end{equation}\begin{equation}
  D_i = \mathit{ReLU}(C^{3\times3}(S_i)), \quad i=1,2,3.
\end{equation}
The final output is produced by a $1\times1$ convolution and sigmoid activation $\sigma$:
\begin{equation}
  I_{out} = \sigma(C^{1\times1}(D_3)).
  \label{eq:output}
\end{equation}

\subsection{Region-Aware Attention (RAA)}\label{subsec:RAAM}
The Region-Aware Attention (RAA)~\cite{naveed2024ra} module, shown in Figure~\ref{LiteFusion-Attention}(a), enhances spatial feature learning before skip connections.

Spatial attention is computed via multi-scale max and average pooling:
\begin{equation}
  m = \mathit{ReLU}(BN(C^{3\times3}(I)))
\end{equation}

\begin{equation}
  m_1 = M_p^{2\times2}M_p^{4\times4}(m), \quad
   m_2 = A_p^{2\times2}A_p^{4\times4}(m), 
\end{equation}

\begin{equation}
  S = m_1 \otimes m_2
\end{equation}

where $M_p$ and $A_p$ denote max and average pooling, respectively, and $\otimes$ is element-wise multiplication. Channel attention is then derived by global averaging:
\begin{equation}
  Att_M = \frac{1}{N} \sum_{i=1}^{N} S_i \left( \frac{1}{M} \sum_{j=1}^{M} m_{i,j} \right),
\end{equation}
and the input is modulated as:
\begin{equation}
  F_{out} = I \otimes Att_M.
\end{equation}

\subsection{LiteFusion Attention Block}\label{subsec:litefusion}
The proposed model seeks to efficiently integrate local and global contexts through LiteFusion Attention. Context aggregation is accomplished by modulation-driven enhancement of attention mechanisms coupled with token mixing. Moreover, it applies channel mixing for adaptive feature refinement, which furthers refinement at the output of each sub-layer. This attention structure employs residual learning to mitigate training instability via gradient passage. Figure \ref{LiteFusion-Attention}(b) shows the LiteFusion attention architecture.


Let $F \in \mathbb{R}^{H \times W \times C}$ be the output feature map with spatial dimensions $H$ and $W$ and $C$ channels from the encoder. The LiteFusion block first computes the global context from $F$ a sequence of $1 \times 1$ and $3 \times 3$ convolutions followed by global average and max-average pooling. The aim of LiteFusion attention blocks is to capture both types of contextual dependencies, global and local, while remaining lightweight. Each block consists of a modulation sub-network, a token mixer, and a channel mixer, all coupled with residual learning to improve gradient flow. 

To generate an intermediate representation, the input feature map (\( f_{map} \)) first goes through sequential $1 \times 1$ and $3 \times 3$ convolution operations followed by layer normalization.

\begin{equation}
    L_1= C^{3\times 3}(\text{LayerNorm}(C^{1\times 1}(f_{map})))
    \label{eq14}
\end{equation}
After that, a $3 \times 3$ convolution, ReLU activation, and global average pooling are applied to calculate a global context vector. The attention weights are generated by passing the resulting feature map through  $1 \times 1$ convolution and a sigmoid activation. The procedure is as follows:

\begin{equation}
    L_2= \sigma(C^{1\times1}(\text{GAP}(\text{ReLU}(C^{1\times 1}(\text{ReLU}(C^{3\times 3}(L_1))))))
    \label{eq15}
\end{equation}

At the same time, a different $3 \times 3$ convolution is used on the intermediate representation to yield spatially filtered features:

\begin{equation}
    L_3= C^{3\times3}(L_1)
    \label{eq16}
\end{equation}

The feature maps are then modulated using the attention weights:
\begin{equation}
    L_4= L_2 \otimes L_3
    \label{eq17}
\end{equation}

A focal modulation operation is subsequently introduced. It initiates by calculating the difference between global max pooling and average pooling of \( L_4 \), adjusted by a modulation hyperparameter (\( \alpha \)=0.25):

\begin{equation}
    m = (\text{MP}(L_4) - \text{AP}(L_4)) \cdot \alpha
    \label{eq18}
\end{equation}

This difference vector undergoes a $1 \times 1$ convolution with a sigmoid activation function applied afterwards in order to produce the modulation vector.

\begin{equation}
    m^{'} = \sigma(C^{1\times1}(m))
    \label{eq19}
\end{equation}

The modulation is applied channel-wise to the previously computed features:
\begin{equation}
    M = L_4 \otimes m^{'}
    \label{eq20}
\end{equation}

To accentuate the prominent features, a non-linear focal enhancement is performed with the help of a power operator, specifically:

\begin{equation}
    M_{\text{out}} = M^{\gamma}
    \label{eq21}
\end{equation}

The value of the focal exponent \( \gamma \) is 2. This enhanced feature map is then fused with the previously modulated output via element-wise multiplication:

\begin{equation}
    L_5 = F_{\text{mod}} \otimes L_4
    \label{eq22}
\end{equation}

A residual projection is used to align and sum feature maps:
\begin{equation}
    L_6 = C^{1\times1}(L_5) + C^{1\times1}(F_{\text{mod}})
    \label{eq23}
\end{equation}

The fused features undergo processing via a token mixer, which consists of layer normalization, depthwise $1 \times 1$ convolution, GELU activation, and dropout:

\begin{equation}
    F_{\text{tok}} = D_r^{0.5}(\text{GELU}(\text{DWC}^{1\times1}(\text{LayerNorm}(L_6))))
    \label{eq24}
\end{equation}

A residual connection adds the original \( L_6 \) back to the token-mixed representation:
\begin{equation}
    F_{\text{tok}}^{'} =  F_{\text{tok}} + \text{Res}(L_6)
    \label{eq25}
\end{equation}

The output is then passed through the channel mixer, which consists of dense transformations with ReLU and dropout for cross-channel interaction:
\begin{equation}
    F_{\text{chan}} = D_r^{0.5}(\text{Dense}(D_r^{0.5}(\text{ReLU}(\text{Dense}(\text{LayerNorm}(F_{\text{tok}}^{'}))))))
    \label{eq26}
\end{equation}

Finally, another residual addition combines the channel-mixed output with the token-mixed representation to produce the final attention-enhanced features:
\begin{equation}
    F_{\text{Lite}} = F_{\text{chan}} + \text{Res}(F_{\text{tok}}^{'})
    \label{eq27}
\end{equation}

This architecture efficiently integrates modulation-based attention, spatial feature mixing, and adaptive channel-wise transformations within a residual learning framework. Consequently, it improves the network’s ability to understand and perceive complexity with a low computational complexity; therefore, the architecture is well-suited for dense predictive tasks like vessel segmentation.

\subsection{Loss Function and Optimization}\label{subsec:loss}
A weighted dice loss function is used to train the proposed model to manage class imbalances efficiently, where higher weights are assigned to under-represented classes. The imbalance in segmentation tasks is addressed by this weighting scheme, which highlights minority class contributions during optimization. In retinal vessel segmentation, background pixels are considerably larger than vessel pixels. The loss function restricts vessel misinterpretation by allocating higher weights to vessel classes, enabling the model to enhance sensitivity to thin structures. The dice loss is defined as:

\begin{equation}
\mathcal{L}_d(S, G) = 1 - \sum_{k=1}^{c} w_k \frac{2 \sum_{j=1}^{n} S(k, j) \cdot G(k, j)}{\sum_{j=1}^{n} S(k, j)^2 + \sum_{j=1}^{n} G(k, j)^2 + \xi}
\label{Loss}
\end{equation}

where \(S\) and \(G\) denote the segmented output and ground truth, respectively, and \( w_k \) represents the weight of the \( k \)-th class, \( c \), \( n \) and \( \xi \) are the number of pixels, the number of classes, and a smoothing constant, respectively. The Adam optimizer is used to optimize the model, ensuring strong convergence. We chose the Adam optimizer because it can change its learning rate throughout training, which is very helpful for segmenting small, thin structures like retinal arteries.

\section{Results}

\subsection{Implementation and datasets}
We use TensorFlow and Keras to create our model using an NVIDIA RTX A4000 with 16 GB of GDDR6 VRAM. The datasets used to evaluate the model comprise only a small number of images, which results in overfitting and poor generalization. Data augmentation solved this challenge by rotating the images by 20 degrees to include modifications in acquisition angles and modifying their contrast to include modifications in lighting and image quality. We applied primary augmentation techniques to DRIVE, STARE, and CHASE\_DB datasets, namely CLoDSA\footnote{{https://github.com/joheras/CLoDSA}} and IMGAUG\footnote{{https://github.com/aleju/imgaug}}.
The DRIVE dataset \cite{qureshi2013manually} contains forty retinal images, each with $565\times 584$ pixels of resolution. The STARE dataset \cite{STAREDataset} has 20 color fundus images with a resolution of $700 \times 605$. The CHASE\_DB dataset \cite{7530915} is composed of 28 images, each representing $1024\times 1024$ pixels of resolution. Table \ref{tab:datasets} shows the detailed insight into the datasets. We used 80\% and 20\% of the images for model training and validation from each dataset with a batch size of 8 and an ADAM optimizer with a learning rate of 0.002. 

\begin{table}[H]
\centering
\caption{Overview of datasets and their properties, including the number of training and testing images, total and augmented images, original image resolution, field of view (FOV), and training details, providing better insight into their application.}
\label{tab:datasets}
\adjustbox{max width=0.5\textwidth}{
\begin{tabular}{lccc}
\toprule
\textbf{Property} & \textbf{DRIVE}~\cite{qureshi2013manually} & \textbf{STARE}~\cite{STAREDataset} & \textbf{CHASE\_DB}~\cite{7530915} \\
\midrule
Training Images & 20 & 16 & 20 \\
Testing Images & 20 & 4 & 8 \\
Total Images & 40 & 20 & 28 \\
Augmented Images & 1080 & 1024 & 1080 \\
Resolution (pixels) & 565$\times$584 & 700$\times$605 & 1024$\times$1024 \\
Resized to & 512 & 512 & 512 \\
Field of View (FOV) & 35 & 45 & 45 \\
\bottomrule
\end{tabular}
}
\end{table}

 Table \ref{complex_results} and Table \ref{tab:Vessels} compare the computational complexity and performance of our proposed model, respectively, in comparison with other state-of-the-art models. Results show that LFA-Net performs well in terms of dice and jaccard on DRIVE, STARE and CHASE\_DB. The segmentation performance and low parameter count of the model surpass computationally expensive models. This makes it much easier to deploy the model while maintaining accuracy in real-time or resource-constrained applications.
\begin{table}[H]
\centering
\caption{Comparison of the computational complexity of LFA-Net with other state-of-the-art methods.}
\label{complex_results}

\adjustbox{max width=0.5\textwidth}{
\begin{tabular}{lcccc}
\toprule
\textbf{Model} & \textbf{Param (M)} & \textbf{FLOPs (G)} & \textbf{Size (MB)} \\
\midrule
UNet\cite{ronneberger2015u}         & 7.76  & 96.68   & 29.60 \\
UNet++\cite{zhou2018unet++}         & 9.04  & 238.52  & 34.49 \\
Att.UNet\cite{oktay2018attention}   & 9.25  & 371.68  & 35.33 \\
IterNet\cite{li2020iternet}         & 13.6  & 194.4   & 94.7  \\
GT-DLA\cite{yuan2021multi}          & 26.0  & 473.9   & 2.3   \\
LiViT-Net\cite{tong2024livit}       & 6.9   & 71.1    & 27.6  \\
FS-UNet\cite{jiang2024retinal}      & 0.87  & 47.6    & 3.50  \\
\textbf{LFA-Net}                     & \textbf{0.11} & \textbf{4.46} & \textbf{0.42} \\
\bottomrule
\end{tabular}}
\end{table}

 \begin{table*}[!ht]
\centering
\caption{Performance comparison of LFA-Net with existing  state-of-the-art methods on the DRIVE, STARE, and CHASE\_DB datasets.}
\adjustbox{max width=\textwidth}{
\begin{tabular}{l c cccc ccccc ccccc}
\toprule
\multirow{2}{*}{\textbf{Method}} & \multirow{2}{*}{\textbf{Param (M)}} & \multicolumn{12}{c}{\textbf{Performance Measures in (\%)}} \\
\cmidrule(lr){3-14}
& & \multicolumn{4}{c}{\textbf{DRIVE}} && \multicolumn{4}{c}{\textbf{STARE}} && \multicolumn{4}{c}{\textbf{CHASE\_DB}} \\
\cmidrule(lr){3-6} \cmidrule(lr){8-11} \cmidrule(lr){13-16}
& & \textbf{Dice} & \textbf{J} & \textbf{Sn} & \textbf{Sp} 
&& \textbf{Dice} & \textbf{J} & \textbf{Sn} & \textbf{Sp} 
&& \textbf{Dice} & \textbf{J} & \textbf{Sn} & \textbf{Sp} \\
\midrule
BCD-UNet \cite{azad2019bi} & 20.65 & 82.49 & 69.33 & 79.84 & 98.03 && 82.30 & 68.14 & 78.92 & 98.16 && 79.32 & 67.42 & 77.35 & 98.01 \\
U-Net++ \cite{zhou2018unet++} & 9.04 & 80.60 & 68.27 & 78.40 & 98.00 && 81.40 & 69.02 & 79.02 & 98.36 && 83.49 & 66.88 & 82.83 & 98.21 \\
Att.Unet \cite{oktay2018attention} & 9.25 & 80.39 & 67.21 & 79.06 & 98.31 && 81.06 & 68.39 & 78.04 & \textbf{98.87} && 79.64 & 66.17 & 80.84 & 98.31 \\
U-Net \cite{ronneberger2015u} & 7.76 & 81.41 & 68.64 & 80.57 & 98.33 && 81.18 & 68.56 & 70.50 & 98.84 && 78.98 & 65.26 & 76.50 & \textbf{98.84} \\
MultiResNet \cite{ibtehaz2020multiresunet} & 7.20 & 82.32 & 69.26 & 79.46 & 97.89 && 82.44 & 68.27 & 77.09 & 98.48 && 80.12 & 67.09 & 80.10 & 98.04 \\
SegNet \cite{badrinarayanan2017segnet} & 1.42 & 83.02 & 70.23 & 80.18 & 98.26 && 83.41 & 70.71 & 80.12 & 98.65 && 81.96 & 68.56 & 81.38 & 98.24 \\
IterNet \cite{li2020iternet} & 13.60 & 82.18 & 69.21 & 77.35 & 98.38 && 81.46 & 68.76 & 77.15 & 98.86 && 80.73 & 67.44 & 79.70 & 98.20 \\
OCE-Net \cite{wei2023orientation} & 6.30 & 83.02 & 71.22 & 80.18 & 98.12 && 83.41 & 73.67 & 80.62 & 98.72 && 81.96 & 69.87 & 81.38 & 98.24 \\
TA\_Mamba \cite{shao2025tubular} & 3.5 & 82.48 & 70.21 & \textbf{84.76} & 98.82 && 81.59 & 69.12 & 82.73 & 99.47 && \textbf{81.68} & 68.82 & \textbf{87.28} & 99.01 \\
DCNet \cite{Shang2024} & 1.05 & 82.94 & 71.15 & 82.08 & 98.02 && 82.50 & 71.35 & 82.30 & 96.77 && 83.32 & 72.46 & 82.05 & 98.17 \\
FS-UNet \cite{jiang2024retinal} & 0.87 & 82.46 & 70.19 & 80.71 & \textbf{98.60} && 84.05 & 72.67 & 83.80 & 98.05 && 81.33 & 68.56 & \textbf{83.42} & 98.53 \\
G-Net Light \cite{iqbal2022g} & 0.39 & 82.02 & 69.09 & 81.92 & 98.29 && 82.78 & 69.64 & 81.70 & 98.53 && 80.48 & 67.76 & 82.10 & 98.38 \\
\textbf{LFA-Net} & \textbf{0.11} & \textbf{83.18 }& \textbf{71.24} & 80.56 & 98.09 && \textbf{87.16} & \textbf{77.29} & \textbf{88.13} & 98.30 && \textbf{84.05} & \textbf{72.52} & 82.25 & 98.15 \\
\bottomrule
\end{tabular}
}
\label{tab:Vessels}
\end{table*}

\begin{figure*}[!ht]
  \centering
    \includegraphics[width=\textwidth]{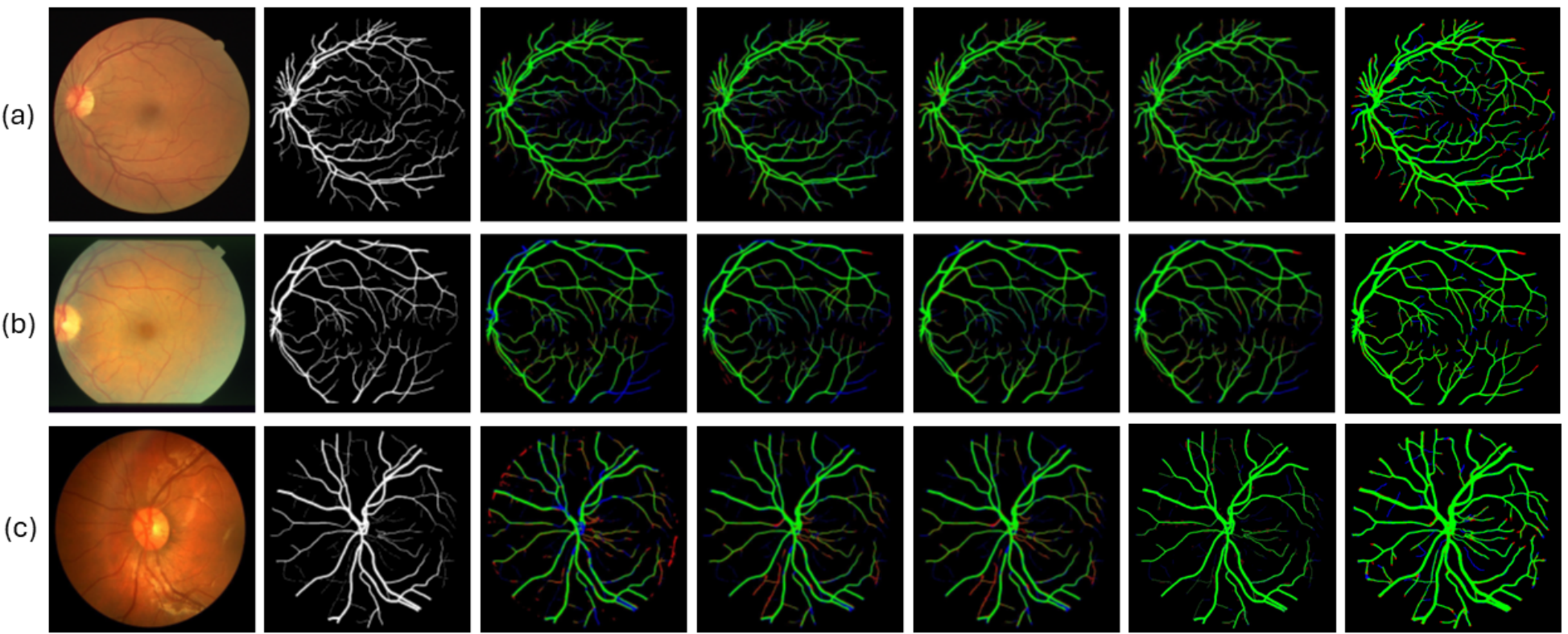}
    \caption{Visual results of retinal vessel segmentation for (a) DRIVE, (b) STARE and (c) CHASE\_DB. The images displayed in the following order are input images, the ground truth and the outputs of the G-Net Light, MultiResNet, SegNet, U-Net++, and LFA-Net models.}\label{fig:combined}
\end{figure*}

Figure \ref{fig:combined} demonstrates the visualization of segmentation results for different existing models across DRIVE, STARE ans CHASE\_DB datasets. Each model's performance is illustrated with color-coded overlays: blue pixels indicate false positives, and green pixels denote true positives. Among the models, LFA-Net yields the most precise segmentations, exhibiting the least amount of false positive and false negative detections.

\subsection{Ablation Study}
\begin{table*}[!ht]
  \centering
  \caption{Ablation study of the proposed model with modifications to evaluate the impact of various components on the baseline model.}
   \resizebox{\textwidth}{!}{%
    \begin{tabular}{lcccccc}
    \toprule
    \multirow{2}[4]{*}{\textbf{Method}} & \multicolumn{6}{c}{\textbf{Performance Measures (\%)}} \\
    \cmidrule{2-7} 
        & \textbf{Param (M)}  & \textbf{Dice} & \textbf{J} & \textbf{Acc} & \textbf{Sen} & \textbf{Sp} \\
    \midrule
       Lightweight UNet no-skip connection (LU-NS) & 0.07 & 80.30 & 64.11 & 93.65 & 69.02 & 96.93 \\
        Multiscale LU no skip connection (MLU-NS) (1\,$\times$\,1, 3\,$\times$\,3, Dilated 3\,$\times$\,3) & 0.08 & 80.92 & 64.01 & 94.62 & 70.10 & 97.10 \\
        Multiscale LU skip connection (MLU) (1\,$\times$\,1, 3\,$\times$\,3, Dilated 3\,$\times$\,3) & 0.09 & 81.32 & 66.99 & 94.66 & 71.47 & 97.43 \\
        MLU + (($\mathcal{R}$)-Skip) & 0.10 & 81.38 & 68.69 & 94.73 & 73.87 & 97.73 \\
         MLU + (($\mathcal{LF}$)-Bottleneck) & 0.10 & 81.61 & 69.11 & 95.07 & 74.48 & 96.57 \\
        MLU + ($\mathcal{R}$)-Skip + ($\mathcal{LF}$)-Bottleneck) & 0.12 & 82.67 & 70.51 & 95.19 & 77.40 & 97.75 \\
         MLU + (($\mathcal{LF}$)-Bottleneck)+(($\mathcal{R}$)-Bottleneck) & 0.10 & 82.61 & 70.11 & 95.27 & 77.18 & 97.67 \\
        MLU + ($\mathcal{R}$) in 1\,3\,-Skip + (($\mathcal{LF}$)-Bottleneck)+(($\mathcal{R}$)-Bottleneck) & 0.11 & 83.01 & 71.09& 95.98 & 79.95 &	98.02\\
        MLU + ($\mathcal{R}$) in 2\,3\,-Skip + (($\mathcal{LF}$)-Bottleneck)+(($\mathcal{R}$)-Bottleneck) & 0.12 & 83.09 & 71.19 & 96.02 & 80.32 & 97.98 \\
       \textbf{MLU + ($\mathcal{R}$) in 1\,2\,-Skip + ($\mathcal{LF}$)-Bottleneck+ ($\mathcal{R}$)-Bottleneck}& 0.11 & \textbf{83.18} & \textbf{71.24} & \textbf{96.09} & \textbf{80.56} & \textbf{98.09} \\
    \bottomrule
    \end{tabular}%
    }
  \label{tab:Ablation}%
\end{table*}%
\begin{figure*}[!ht]
\centering
\includegraphics[width=\textwidth]{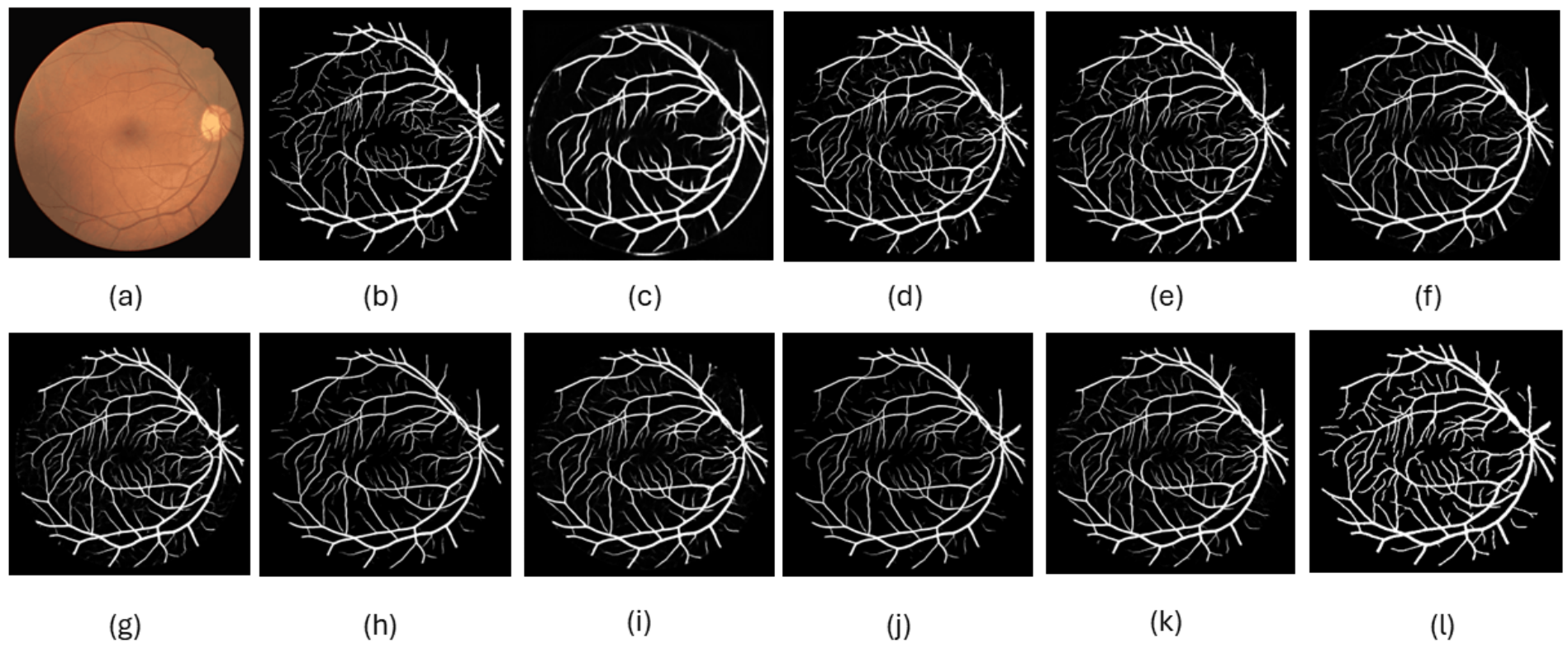}
    \caption{Visual comparison of vessel segmentation results from different architectural configurations used in the ablation study. (a) Input RGB image, (b) Ground truth, (c) Lightweight U-Net without skip connections (LU-NS), (d) Multiscale LU without skip connections (MLU-NS), (e) Multiscale LU with skip connections (MLU), (f) MLU + ($\mathcal{R}$)-Skip, (g) MLU + ($\mathcal{LF}$)-Bottleneck, (h) MLU + ($\mathcal{R}$)-Skip + ($\mathcal{LF}$)-Bottleneck, (i) MLU + ($\mathcal{LF}$)-Bottleneck + ($\mathcal{R}$)-Bottleneck, (j) MLU + ($\mathcal{R}$)-Skip + ($\mathcal{LF}$)-Bottleneck + ($\mathcal{R}$)-Bottleneck, (k) MLU + ($\mathcal{R}$) in 2–3 skip connections + both bottlenecks, and (l) MLU + ($\mathcal{R}$) in 1–2 skip connections + ($\mathcal{LF}$)-Bottleneck + ($\mathcal{R}$)-Bottleneck.}
\label{fig:ablation}
 
\end{figure*}

Table~\ref{tab:Ablation} presents the results of an ablation study that systematically evaluates the contribution of various architectural elements to the baseline model. 
A lightweight UNet without skip connections (LU-NS), which serves as the starting point, achieves a dice score of 80.30\%. Multiscale feature extraction slightly improves dice and specificity when multiscale convolutions without skip connections (MLU-NS) are included. Retaining spatial information is important, as adding skip connections to the MLU backbone (MLU+($\mathcal{R}$)-Skip) significantly increases the dice to 81.38\% and sensitivity to 79.27\%. Incorporating RAA into the bottleneck (($\mathcal{R}$)-Bottleneck) or skip connections tends to enhance all performance measures, particularly when paired with the custom LiteFusion-Attention bottleneck (($\mathcal{LF}$)-Bottleneck). With a dice of 83.18\%, a jaccard of 71.24\%, an accuracy of 96.09\%, a sensitivity of 80.56\%, and a specificity of 98.09\%, the configuration with RAA in 1-2 skip connections and both ($\mathcal{LF}$)- and ($\mathcal{R}$)-bottlenecks achieves the best overall performance. This demonstrates that multiscale context, skip connections, and attention mechanisms produce the most reliable segmentation outcomes.\\

Although the numerical improvements described in the table may appear relatively tiny, they can be clinically important in retinal vascular segmentation. Small increases in dice and sensitivity, in particular, frequently result in more accurate segmentation of thin and low-contrast capillaries, which are critical for detecting microvascular abnormalities in situations such as diabetic retinopathy and dementia. Figure~\ref{fig:ablation} demonstrates that different configurations result in more continuous vascular architectures and fewer false positives in background regions. As the design improves, background noise decreases, thinner branches are accurately identified, and the segmented vessels seem relatively continuous. As a result, even minor metric improvements generate more reliable and diagnostically important segmentation results for practical applications. 

\section{Conclusion}

In this paper, we introduced LFA-Net, a lightweight and efficient retinal vessels segmentation model, achieving a good balance between accuracy and efficiency. The LiteFusion-Attention module is a novel attention mechanism inspired by Vision Mamba and modulation-based attention. It combines adaptive channel scaling, token mixing, residual connection, and modulation-based contextual refinement. LFA-Net can capture fine-grained vessel details while keeping the model lightweight. We evaluate the model on three benchmark datasets—DRIVE, STARE, and CHASE\_DB— and conclude that LFA-Net maintained strong performance with competitive dice and jaccard scores. Its lightweight design emphasizes its efficiency, with parameters of 0.11 million, a memory footprint of 0.42 MB, and a FLOP estimate of 4.46 GFLOP. Overall, the findings indicate that LFA-Net is well-suited for real-time applications and deployment in situations with limited processing resources, making it a promising tool for practical and cost-effective retina image analysis.

\bibliographystyle{IEEEtran} 

\end{document}